%% file: main.tex
\documentclass[10pt,twocolumn,letterpaper]{article}

\usepackage[pagenumbers]{cvpr} % To force page numbers, e.g. for an arXiv version
\usepackage[accsupp]{axessibility}  % Improves PDF readability for those with disabilities.

\usepackage{amsmath}
\usepackage{algorithm}
\usepackage{algorithmic}
\usepackage{multirow}
\usepackage{booktabs}
\usepackage{amssymb}
\usepackage{enumitem}
\usepackage{color}
\usepackage{xcolor}

\input{preamble}
\definecolor{cvprblue}{rgb}{0.21,0.49,0.74}
\usepackage[pagebackref,breaklinks,colorlinks,allcolors=cvprblue]{hyperref}

\def\paperID{14741} % *** Enter the Paper ID here
\def\confName{CVPR}
\def\confYear{2026}

\title{D4C: Data-Free Quantization for Contrastive  Language-Image \\ Pre-training Models}

\author{
Wenlun Zhang\textsuperscript{1} \quad
Yunshan Zhong\textsuperscript{2\thanks{Corresponding author.}} \quad
Zihao Ding\textsuperscript{1} \quad
Xinyu Li\textsuperscript{1} \quad
Kentaro Yoshioka\textsuperscript{1} \\
\textsuperscript{1}Department of Electronics and Electrical Engineering, Keio University \\
\textsuperscript{2}School of Computer Science and Technology, Hainan University \\
{\tt\small \{wenlun\_zhang, kyoshioka47\}@keio.jp} \quad {\tt\small viperzhong@163.com}
}

\begin{document}
\maketitle
\input{sec/0_Abstract}    
\input{sec/1_Introduction}
\input{sec/2_Related_Works}
\input{sec/3_Methodology}
\input{sec/4_Experiments}
\input{sec/5_Limitations}
\input{sec/6_Conclusion}

\section*{Acknowledgments}

This research was supported in part by JST CREST JPMJCR23M4 and JPMJCR21D2, including AIP challenge program, JST Next-generation Edge AI Semiconductor JPMJES2515, JST SPRING JPMJSP2123, and JSPS Kakenhi 23H00467, 24K02940.

{
    \small
    \bibliographystyle{ieeenat_fullname}
    \bibliography{main}
}


\end{document}

%% file: sec/0_Abstract.tex
\begin{abstract}
Data-Free Quantization (DFQ) offers a practical solution for model compression without requiring access to real data, making it particularly attractive in privacy-sensitive scenarios. While DFQ has shown promise for unimodal models, its extension to Vision-Language Models such as Contrastive Language-Image Pre-training (CLIP) models remains underexplored. In this work, we reveal that directly applying existing DFQ techniques to CLIP results in substantial performance degradation due to two key limitations: insufficient semantic content and low intra-image diversity in synthesized samples. To tackle these challenges, we propose D4C, the first DFQ framework tailored for CLIP. D4C synthesizes semantically rich and structurally diverse pseudo images through three key components: \textbf{1)} Prompt-Guided Semantic Injection aligns generated images with real-world semantics using text prompts; \textbf{2)} Structural Contrastive Generation reproduces compositional structures of natural images by leveraging foreground-background contrastive synthesis; and \textbf{3)} Perturbation-Aware Enhancement applies controlled perturbations to improve sample diversity and robustness. These components jointly empower D4C to synthesize images that are both semantically informative and structurally diverse, effectively bridging the performance gap of DFQ on CLIP. Extensive experiments validate the effectiveness of D4C, showing significant performance improvements on various bit-widths and models. \textbf{Code is available at} \href{https://github.com/Keio-CSG/D4C}{\textit{https://github.com/Keio-CSG/D4C}}
\end{abstract}

%% file: sec/1_Introduction.tex
\section{Introduction}

Vision-Language Models (VLMs), such as Contrastive Language-Image Pre-training (CLIP) models~\cite{CLIP}, which integrate visual and textual modalities, have propelled progress across a range of fields, including healthcare diagnostics~\cite{CLIP_Medical}, autonomous driving~\cite{CLIP-RLDrive}, and medical image analysis~\cite{MedCLIP,MedCLIP-SAM}. Despite their impressive performance, these models often demand substantial computational and memory resources, limiting their practicality in resource-constrained environments. To address this issue, model quantization~\cite{White_Paper} has emerged as an effective technique, compressing models by transforming floating-point parameters into low-bit representations, thereby reducing both memory footprint and inference latency.

Conventional quantization methods require access to training data, which may not be viable in privacy-sensitive applications. A representative example is CLIP models trained on medical text-image datasets~\cite{PathGen,MedCLIP}, where the data itself may be highly confidential. To overcome this limitation, Data-Free Quantization (DFQ) provides a promising alternative by generating pseudo samples directly from pre-trained models~\cite{GenerativeDFQ,GENIE,Sharpness-Aware_Gen,SMI,TexQ,UDFC}, which are then used for model calibration. Over the past few years, DFQ has achieved notable success in both Convolutional Neural Networks (CNNs) and Vision Transformers (ViTs). In CNNs, DFQ typically uses Batch Normalization Statistics (BNS) to guide the image generation process~\cite{ZeroQ}, whereas ViT-based approaches exploit the inherent attention mechanisms to produce more diverse and informative synthetic samples~\cite{PSAQ}. However, existing DFQ frameworks are designed for unimodal architectures, and little attention has been paid to extending DFQ to cross-modal models like CLIP, leaving a critical research gap unaddressed.

\begin{figure}[htbp]
    \centering
    \includegraphics[width=\columnwidth]{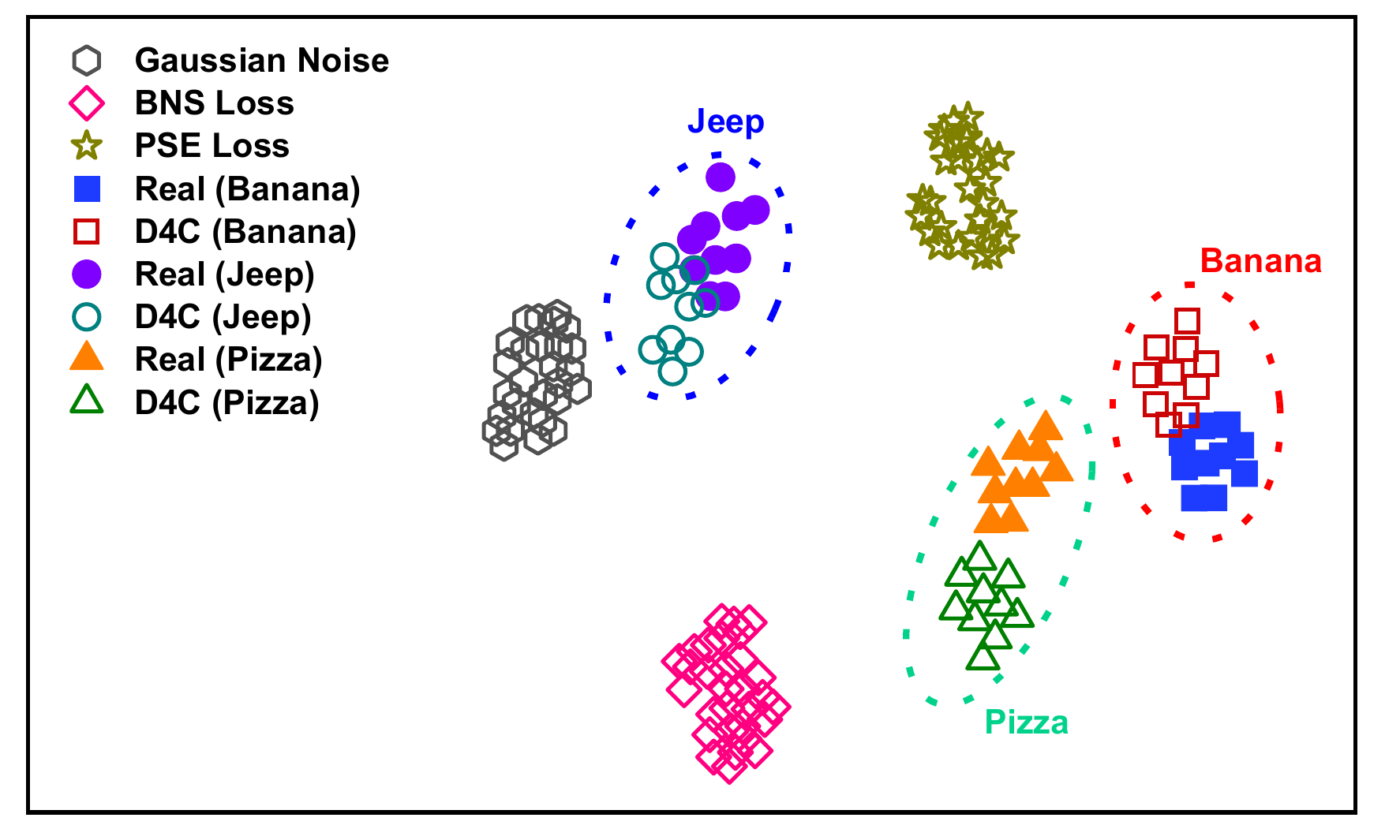}
    \caption{UMAP~\cite{UMAP} visualization of features across various samples. Images generated using BNS or PSE losses are distant from real images, indicating limited semantic information. In contrast, D4C-generated samples closely match real data.}
    \label{Fig_Feature_UMAP}
\end{figure}

\begin{figure}[htbp]
    \centering
    \includegraphics[width=\columnwidth]{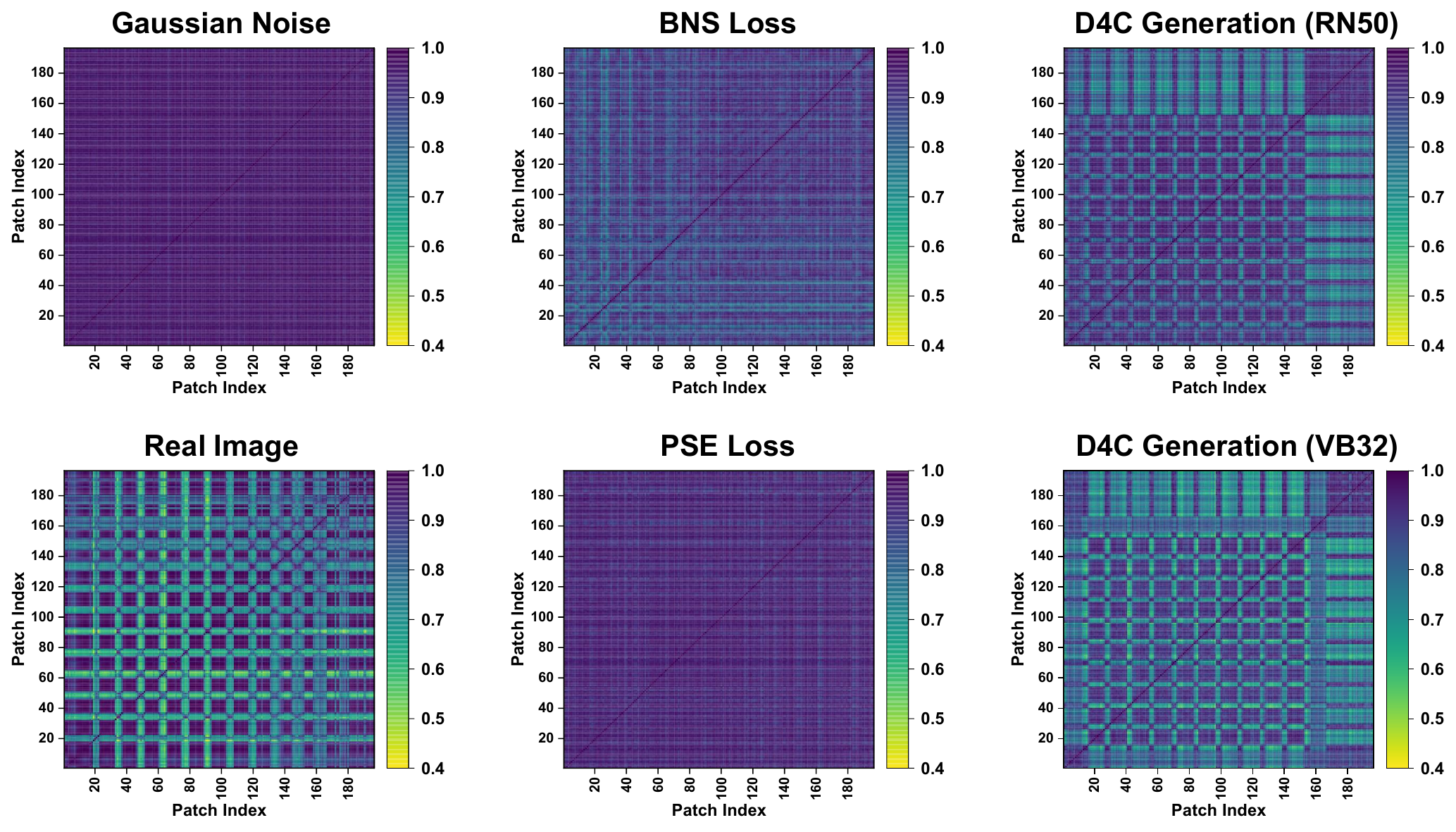}
    \caption{Patch similarity visualization across different images. Compared to Gaussian noise and BNS/PSE-based synthetic images, which exhibit weak or irregular internal patch relationships, D4C-generated images display structured similarity patterns closely resembling those of real images.}
    \label{Fig_Patch_Similarity}
\end{figure}

In this paper, we observe that directly applying existing DFQ methods to CLIP leads to a substantial performance degradation compared to using real data, as demonstrated in Table~\ref{Table_Exp_Result}. To better understand this issue, we identify two key challenges in synthesizing pseudo data for CLIP quantization. First, current DFQ techniques often generate synthetic images with insufficient semantic content, resulting in poor alignment in the latent feature space. As illustrated in Fig.~\ref{Fig_Feature_UMAP}, pseudo samples produced by conventional methods tend to collapse into indistinct clusters, indicating limited semantic expressiveness. Second, these synthetic samples exhibit low intra-image diversity, failing to reflect the structural complexity of natural images. For instance, Fig.~\ref{Fig_Patch_Similarity} shows that existing methods produce overly uniform similarity patterns, which hinders their effectiveness for quantization. Given that CLIP is trained using large-scale image-text pairs, effective quantization requires calibration samples that preserve both semantic content and structural complexity. Without such qualities, the quantization process suffers a significant loss in accuracy.

To address the aforementioned challenges, we propose a novel \underline{D}FQ Framework \underline{For} \underline{C}LIP, dubbed D4C. First, we introduce Prompt-Guided Semantic Injection (PGSI), which injects high-level semantics into pseudo images by guiding the synthesis process with text prompts, thereby improving their latent alignment with real samples. To mitigate intra-image homogeneity, we propose Structural Contrastive Generation (SCG), which leverages foreground-background contrastive synthesis to reproduce the compositional structures of natural images. Additionally, Perturbation-Aware Enhancement (PAE) applies controlled perturbations during generation to increase sample diversity and robustness. Together, these components enable D4C to synthesize data that are both semantically informative and structurally realistic, significantly narrowing the DFQ performance gap on CLIP. Our main contributions are as follows:

\begin{itemize}[leftmargin=*]
    \item We identify and analyze the limitations of directly applying existing DFQ methods to CLIP, revealing two critical challenges: semantic insufficiency and structural homogeneity in the generated pseudo data.
    \item We propose D4C, the first DFQ framework for CLIP that integrates PGSI, SCG, and PAE to jointly enhance the semantic expressiveness and structural diversity of synthetic samples, thereby significantly improving their effectiveness for CLIP quantization.
    \item We conduct extensive experiments to validate the effectiveness of D4C. The results consistently demonstrate that our method outperforms existing DFQ techniques and, to the best of our knowledge, establishes the first strong baseline for DFQ on CLIP. Specifically, under W4A8 quantization for RN50 and VB32\footnote{We denote ResNet-50, ResNet-50x16, ViT-B/32, and ViT-B/16 as RN50, RN50x16, VB32, and VB16 in this paper, respectively.} in zero-shot classification tasks, D4C achieves accuracy improvements of 12.4/18.9\%, 6.8/19.7\%, and 1.4/5.7\% over DFQ baselines on CIFAR-10, CIFAR-100, and ImageNet-1K, respectively.
\end{itemize}

%% file: sec/2_Related_Works.tex
\section{Related Works}

\subsection{Contrastive Language-Image Pre-training Models}

CLIP~\cite{CLIP} marks a significant advancement in vision-language modeling by leveraging large-scale web data and contrastive learning to directly align image and text embeddings. This approach not only enhances its generalization capabilities but also provides superior zero-shot transfer performance across diverse tasks, including image classification, retrieval, and captioning. Driven by CLIP’s success, several domain-specific variants have emerged, notably MedCLIP~\cite{MedCLIP}, which facilitates representation learning from unpaired medical image-text datasets, and adaptations such as MedCLIP-SAM~\cite{MedCLIP-SAM}, aimed at universal medical image segmentation. However, quantizing CLIP in such privacy-sensitive domains is challenging due to the lack of calibration data, motivating our investigation of DFQ for CLIP-based VLMs.

\subsection{Model Quantization}

\subsubsection{Data-Driven Quantization}

Data-driven quantization approaches typically include Quantization-Aware Training (QAT) and Post-Training Quantization (PTQ). QAT~\cite{LSQ,DSQ,APoT,Osc-Free_Quant} achieves high quantization accuracy by incorporating quantization constraints directly into the training phase, but this strategy requires full access to the original training dataset and extensive retraining, imposing substantial computational overhead. Consequently, its practicality is often restricted, especially for large-scale models or datasets. PTQ~\cite{PTQ,AdaRound,BRECQ,QDrop,P4Q,PD-Quant,LiDAR-PTQ,Accurate_PTQ_ViT,ERQ} addresses these limitations by directly quantizing pre-trained models without full retraining, usually relying on a limited subset of training samples. This significantly reduces computational costs and enhances applicability to resource-constrained deployment scenarios~\cite{AHCPTQ}. Nevertheless, even minimal access to real calibration data can be problematic or impossible in sensitive domains such as healthcare, finance, or security, where data privacy and confidentiality constraints are prevalent.

\subsubsection{Data-Free Quantization}

DFQ has emerged as a practical alternative to conventional data-driven approaches, aiming to eliminate dependence on real calibration data entirely. DFQ leverages the pre-trained model itself to generate synthetic data for quantization purposes. For CNNs, ZeroQ~\cite{ZeroQ} leverages BNS from pre-trained models to synthesize calibration samples for quantization. DSG~\cite{DSG} improves upon ZeroQ by introducing slack distribution alignment and layer-wise sample enhancement to mitigate sample homogenization. IntraQ~\cite{IntraQ} further extends this approach by incorporating class-conditional generation using category labels, achieving enhanced quantization performance. For ViTs, PSAQ-ViT~\cite{PSAQ,PSAQ-V2} synthesizes calibration samples by exploiting patch similarity entropy (PSE) derived from self-attention layers. MimiQ~\cite{MimiQ} enhances DFQ performance by aligning the multi-head attention maps and applying structured head-wise distillation to generate consistent pseudo images. SARDFQ~\cite{SARDFQ} further improves image quality by employing attention priors alignment, and multi-semantic reinforcement strategies, effectively mitigating semantic distortions during image synthesis. SynQ~\cite{KimKK25} introduces a novel framework for DFQ by integrating a low-pass filter, class activation map alignment, and difficult samples learning. \cite{KimCLCK25} presents a more comprehensive DFQ survey, which may be beneficial for further research. Although DFQ techniques have demonstrated effectiveness for both CNNs and ViTs, their extension to multimodal architectures like CLIP remains limited, highlighting a critical challenge for the research community.

%% file: sec/3_Methodology.tex
\section{Methodology}

\subsection{Preliminaries}

\subsubsection{Quantization}

Quantization maps floating-point values to discrete integer representations, significantly reducing computational and memory overhead. Formally, the quantization and dequantization processes of the popular uniform quantization can be described as:
\begin{equation}
x_{q} = \text{clamp} \left( \left\lfloor \frac{x}{s} \right\rceil + z, 0, 2^k - 1 \right),
\label{Eq_Quant}
\end{equation}
\begin{equation}
x \approx \hat{x} = s \cdot (x_{q} - z),
\label{Eq_Dequant}
\end{equation}
\noindent where $s$ and $z$ are the scale and zero-point parameters, respectively, and $k$ denotes the quantization bit-width.

\subsubsection{Optimization-Guided PTQ}

In this paper, we adopt a widely used optimization-guided PTQ method during the quantization phase of DFQ. Specifically, we perform block- or layer-wise reconstruction by minimizing the mean squared error between the original and quantized outputs:
\begin{equation}
\begin{aligned}
  \mathcal{L} & =  \| \mathbf{O}_{i} - \hat{\mathbf{O}}_{i} \|_2^2,
  \label{Eq_Recon}
\end{aligned}
\end{equation}
where $ \mathbf{O}_{i}$ and $ \hat{\mathbf{O}}_{i} $ denote the floating-point and quantized outputs of the $i$-th block or layer, respectively.

\subsubsection{Data Synthesis}

DFQ generates synthetic data directly from the pre-trained model without relying on real samples. This process begins with Gaussian noise inputs $\tilde{S} \sim \mathcal{N}(0, 1)$, which are optimized using a synthesis loss function $\mathcal{L}_{\text{Syn}}$. The goal is to obtain pseudo data $S$ that can effectively support model quantization. The effectiveness of DFQ largely depends on the formulation of $\mathcal{L}_{\text{Syn}}$, which typically distills prior knowledge from the model. Notable examples include the BNS loss~\cite{ZeroQ} for CNNs and the PSE loss~\cite{PSAQ} for ViTs.

\begin{figure*}[htbp]
    \centering
    \includegraphics[width=\textwidth]{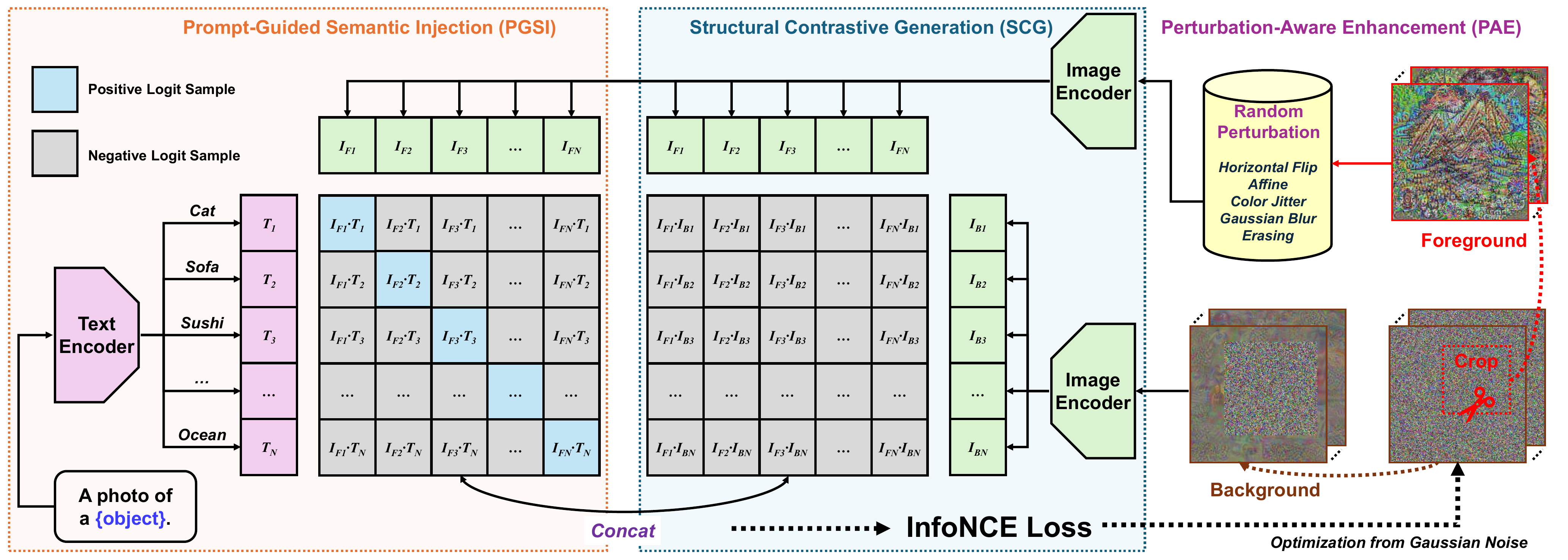}
    \caption{Overview of D4C: PGSI injects semantic information into synthetic samples through object concept prompting; SCG enhances structural diversity via foreground-background contrastive generation; and PAE introduces perturbations to further improve sample quality and expressiveness.}
    \label{Fig_Framework}
\end{figure*}

\subsection{DFQ Challenges of CLIP}

Although BNS and PSE losses have proven effective for DFQ in CNNs and ViTs, they are designed for unimodal models and fall short when applied to CLIP, suffering from considerable performance drops. As shown in Table~\ref{Table_Exp_Result}, W4A8 quantization of CLIP on CIFAR-10 yields only 48.9\% with BNS (RN50) and 53.7\% with PSE (VB32), compared to 71.7\% and 89.6\% when using real data. These stark gaps underscore the limitations of existing DFQ techniques on CLIP and drive our effort to explore solutions tailored to their unique challenges.

The \textbf{first challenge} arises from the observation that synthetic images often lack meaningful semantic content, deviating from real image distributions and resulting in suboptimal quantization performance. To examine this issue, we extract features using CLIP's VB32 encoder from real images, Gaussian noise, and synthetic samples generated via BNS and PSE losses using RN50 and VB32 encoders. These features are projected into a 2D space using Uniform Manifold Approximation and Projection (UMAP)~\cite{UMAP}. As illustrated in Fig.~\ref{Fig_Feature_UMAP}, synthetic samples generated with BNS and PSE cluster tightly, resembling Gaussian noise, whereas real images form three clearly separated semantic groups (\emph{i.e.}, banana, jeep, and pizza). Given that CLIP is pre-trained on large-scale image-text pairs and strongly depends on semantic alignment between modalities, calibration data with rich semantics is crucial for effective quantization. However, synthetic samples produced by conventional DFQ methods show limited semantic diversity in the latent space, making them suboptimal for CLIP. Such semantically deficient pseudo data tend to cause overfitting to quantization parameters, ultimately leading to performance degradation.

The \textbf{second challenge} lies in the lack of structural diversity within synthetic images, as their internal patch relationships are overly homogeneous and deviate significantly from the distribution patterns observed in real data. To demonstrate this, we randomly select samples, divide each image into $16\times16$ patches, resize each patch to $224\times224$, and extract their features using a pre-trained ResNet-18 model~\cite{ResNet}. We then compute the pairwise cosine similarity between patches. As visualized in Fig.~\ref{Fig_Patch_Similarity}, real images exhibit structured and periodic similarity patterns, where alternating high- and low-similarity regions reflect the coexistence of semantically related and unrelated content, such as foreground objects and background context. In contrast, synthetic images generated by BNS or PSE losses fail to capture this structural diversity and display weak or overly uniform similarity patterns, closely resembling Gaussian noise. The resulting lack of intra-sample diversity negatively impacts the quality of synthetic data for quantization.

\subsection{D4C}

To address the aforementioned challenges, we propose D4C, a novel framework specifically designed for the DFQ scenario on CLIP. As shown in Fig.~\ref{Fig_Framework}, D4C comprises three core components: PGSI, which injects semantic information into synthetic data by prompting diverse object concepts; SCG, which enhances structural diversity through foreground-background contrastive generation; and PAE, which further enriches image quality and expressiveness by introducing controlled perturbations during generation.

\subsubsection{Prompt-Guided Semantic Injection}

To address the challenge of synthetic images lacking semantic content, we introduce PGSI, which utilizes text prompts to embed semantic information into the generated images. As shown in the left panel of Fig.~\ref{Fig_Framework}, PGSI constructs a batch of textual prompts based on selected object categories and encodes them using the text encoder. For text inputs, we use the template ``A photo of \{c\}'' with a broad category set (animals, objects, food, scenes) for concept coverage. Simultaneously, a batch of images, initialized from Gaussian noise, is processed by the image encoder. PGSI then applies the InfoNCE loss~\cite{InfoNCE} to encourage strong alignment between matched image-text pairs ($I_i \cdot T_i$) while penalizing mismatched pairs:
\begin{equation}
\mathcal{L}_{\text{InfoNCE}} = -\frac{1}{N} \sum_{i=1}^{N} \log \frac{\exp(I_i \cdot T_i / \tau)}{\sum\limits_{j=1}^{N} \exp(I_i \cdot T_j / \tau)},
\label{Eq_InfoNCE}
\end{equation}
where $I_i$ and $T_i$ represent the $i$-th image and text embeddings, $N$ is the batch size, and $\tau$ is a temperature parameter. This contrastive objective guides the optimization of synthetic images, ensuring that each sample semantically aligns with a meaningful real-world concept.

To validate the effectiveness of the proposed PGSI, we randomly sample generated images from three categories (banana, jeep, and pizza) and extract their features using the VB32 encoder. The projected latent features are visualized in Fig.~\ref{Fig_Feature_UMAP}. As illustrated, the features of the generated samples form three clearly separated clusters, closely aligned with those of real images. This alignment indicates strong semantic consistency between the synthetic and real data, highlighting the semantic enrichment achieved through PGSI. Consequently, the improved semantic information of the synthetic samples contributes to enhanced quantization performance.

\subsubsection{Structural Contrastive Generation}

To address the second challenge that corresponds to insufficient structural diversity in synthetic images, we propose SCG, which promotes the generation of semantically diverse regions within each image. Natural images typically comprise multiple regions, including a foreground containing the target object and a background composed of semantically unrelated content. The similarity between foreground and background regions is generally lower than that within each region. To emulate this property, SCG introduces a region-aware contrastive strategy. Specifically, for each synthetic image, we randomly initialize a foreground bounding box, crop and resize the corresponding region, and extract the foreground embedding $I_{F}$, as illustrated in the middle panel of Fig.~\ref{Fig_Framework}. Simultaneously, we mask the foreground region to obtain the background embedding $I_{B}$. We then compute two similarity matrices: \textbf{1)} between the foreground embeddings $I_F$ and text prompts $T$, and \textbf{2)} between $I_F$ and background embeddings $I_B$. These matrices are concatenated to form a single pool of negative and positive samples within the InfoNCE loss. Noted that these two InfoNCE losses can be unified with the original InfoNCE loss (Eq.\ref{Eq_InfoNCE}) in PGSI, which results in the following SCG losses:
\begin{equation}
\mathcal{L}_{\text{SCG}} = -\frac{1}{N} \sum_{i=1}^{N} \log \frac{\exp(I_{Fi} \cdot T_i / \tau)}{\sum\limits_{j=1}^{N} \exp(\texttt{Cat}(I_{Fi} \cdot T_j, I_{Fi} \cdot I_{Bj}) / \tau)},
\label{Eq_SCG}
\end{equation}
where only the foreground embedding $I_{Fi} \cdot T_i$ serves as the positive image-text pair, while all mismatched image-text similarities $I_{Fi} \cdot T_j$ and foreground-background similarities $I_{Fi} \cdot I_{Bj}$ are included in the denominator.  This unified design facilitates both effective representation learning and efficient backpropagation. By introducing two types of negative pairs, SCG loss not only encourages alignment between the foreground and its assigned prompt but also explicitly pushes it away from irrelevant prompts and their corresponding background. This contrastive setup promotes diverse internal structure within synthetic images and leads to improved synthesis quality.

To validate the effectiveness of SCG, we extract patch features from synthetic images generated by RN50 and VB32 encoders and perform the same patch similarity analysis. As shown in Fig.~\ref{Fig_Patch_Similarity}, the resulting similarity maps exhibit a staggered pattern, reflecting alternating foreground and background regions that closely resemble those in real images. This structural alignment indicates that the generated samples possess richer semantic and spatial information, leading to improved consistency with real data and enhanced quantization performance.

\begin{algorithm}[!t]
\caption{D4C DFQ Pipeline}
\label{Algorithm_D4C}
\textbf{Input}: A pre-trained floating-point CLIP model $\mathcal{F}$ consisting of image encoder $f_I$ and text encoder $f_T$.
\\
\textbf{Output}: Quantized CLIP model $\mathcal{F}_Q$. \\
\textbf{\textcolor{teal}{\# Stage 1: Sample Synthesis}}
\begin{algorithmic}[1] %[1] enables line numbers
\STATE Initialize Gaussian noise images $\tilde{S}$ and foreground bounding boxes $\mathcal{C}$.
\WHILE{not converged}
\STATE Crop foreground regions from $\tilde{S}$ based on $\mathcal{C}$ and resize them.
\STATE Apply PAE to foreground patches.
\STATE Generate background images by masking $\mathcal{C}$ in $\tilde{S}$ with Gaussian noise.
\STATE Encode foreground features $I_{Fi} = f_I(\cdot)$, background features $I_{Bi} = f_I(\cdot)$, and text prompts $T_j = f_T(\cdot)$.
\STATE Compute contrastive logits and obtain final synthesis loss $\mathcal{L}_{\text{Syn}}$ following Eq.~\ref{Eq_Syn_Objective}.
\STATE Update synthetic images $\tilde{S}$ via gradient descent.
\ENDWHILE
\end{algorithmic}
\textbf{\textcolor{teal}{\# Stage 2: Model Quantization}}
\begin{algorithmic}[1]
\STATE Obtain final synthetic images $S$ for image encoder, and reuse PGSI prompts for text encoder calibration.
\FOR{$i = 1, 2, \cdots, N$-th block/layer in image encoder $f_I$ and text encoder $f_T$}
\STATE Conduct reconstruction following Eq.~\ref{Eq_Recon}.
\ENDFOR
\end{algorithmic}
\end{algorithm}

\begin{table*}[t]
\small
\centering
\caption{Quantization results of D4C for zero-shot classification across four types of image encoders. Top-1 accuracy (\%) is reported on CIFAR-10, CIFAR-100, and ImageNet-1K. "W/A" indicates the bit-width of weights and activations. D4C consistently outperforms existing BNS and PSE approaches, achieving state-of-the-art performance.}
\label{Table_Exp_Result}
\begin{tabular}{c|c|c|ccc|c|ccc|c|ccc}
\toprule
\multirow{2}{*}{\textbf{Model}} & \multirow{2}{*}{\textbf{Method}} & \multicolumn{4}{c|}{\textbf{CIFAR-10}} & \multicolumn{4}{c|}{\textbf{CIFAR-100}} & \multicolumn{4}{c}{\textbf{ImageNet-1K}} \\
\cmidrule(lr){3-6} \cmidrule(lr){7-10} \cmidrule(lr){11-14}
& & \textbf{FP} & \textbf{W4A8} & \textbf{W6A6} & \textbf{W8A8} & \textbf{FP} & \textbf{W4A8} & \textbf{W6A6} & \textbf{W8A8} & \textbf{FP} & \textbf{W4A8} & \textbf{W6A6} & \textbf{W8A8} \\
\midrule
\multirow{4}{*}{\textbf{RN50}} 
& Real & \multirow{4}{*}{71.5} & 71.7 & 73.8 & 73.0 & \multirow{4}{*}{40.3} & 38.9 & 37.6 & 39.0 & \multirow{4}{*}{59.8} & 44.9 & 45.5 & 49.7 \\
& Gaussian & & 34.8 & 58.9 & 69.9 & & 8.0 & 25.7 & 38.5 & & 26.4 & 39.5 & 49.5 \\
& BNS\&PSE & & 48.9 & 66.9 & 69.8 & & 20.1 & 31.6 & 38.6 & & 42.9 & 44.0 & 49.1 \\
& D4C & & \textbf{61.3} & \textbf{73.4} & \textbf{70.4} & & \textbf{26.9} & \textbf{36.4} & \textbf{40.8} & & \textbf{44.3} & \textbf{47.9} & \textbf{51.1} \\
\midrule
& Real & \multirow{4}{*}{81.4} & 82.3 & 82.1 & 81.8 & \multirow{4}{*}{52.2} & 48.6 & 48.7 & 50.4 & \multirow{4}{*}{70.7} & 60.8 & 59.0 & 62.8 \\
\textbf{RN50} & Gaussian & & 74.6 & 78.3 & 80.0 & & 43.2 & 44.4 & 49.4 & & 55.0 & 55.2 & 62.0 \\
\textbf{x16} & BNS\&PSE & & 76.4 & 78.0 & 80.2 & & 45.9 & 48.0 & 50.1 & & 58.5 & 58.6 & 63.0 \\
& D4C & & \textbf{77.9} & \textbf{78.7} & \textbf{80.4} & & \textbf{46.2} & \textbf{48.2} & \textbf{50.3} & & \textbf{60.4} & \textbf{58.7} & \textbf{64.8} \\
\midrule
\multirow{4}{*}{\textbf{VB32}} 
& Real & \multirow{4}{*}{89.8} & 89.6 & 89.3 & 90.0 & \multirow{4}{*}{64.2} & 59.8 & 59.2 & 62.1 & \multirow{4}{*}{63.3} & 47.6 & 48.1 & 50.4 \\
& Gaussian & & 18.0 & 19.1 & 24.5 & & 2.8 & 3.0 & 4.8 & & 13.6 & 15.7 & 27.4 \\
& BNS\&PSE & & 53.7 & 57.7 & 68.1 & & 22.1 & 24.9 & 35.5 & & 40.4 & 42.9 & 47.6 \\
& D4C & & \textbf{72.6} & \textbf{75.5} & \textbf{80.4} & & \textbf{41.8} & \textbf{45.8} & \textbf{51.0} & & \textbf{46.1} & \textbf{46.1} & \textbf{51.7} \\
\midrule
\multirow{4}{*}{\textbf{VB16}} 
& Real & \multirow{4}{*}{90.8} & 91.8 & 85.9 & 91.0 & \multirow{4}{*}{66.9} & 63.4 & 48.0 & 64.7 & \multirow{4}{*}{68.3} & 52.7 & 45.3 & 56.4 \\
& Gaussian & & 24.5 & 15.8 & 72.3 & & 5.7 & 2.4 & 39.4 & & 15.3 & 0.8 & 49.0 \\
& BNS\&PSE & & 70.6 & 15.4 & 81.6 & & 39.1 & 4.9 & 50.8 & & 47.0 & 10.5 & 56.1 \\
& D4C & & \textbf{81.4} & \textbf{48.3} & \textbf{86.5} & & \textbf{49.8} & \textbf{19.5} & \textbf{59.7} & & \textbf{49.9} & \textbf{35.7} & \textbf{57.9} \\
\bottomrule
\end{tabular}
\end{table*}

\subsubsection{Perturbation-Aware Enhancement}

While PGSI and SCG collaboratively guide the generation of semantically meaningful and structurally diverse images, relying solely on these components can lead to overfitting during optimization, ultimately limiting sample quality~\cite{TexQ,IntraQ}. In order to improve both diversity and robustness, we incorporate PAE into the generation pipeline. As illustrated in the right panel of Fig.~\ref{Fig_Framework}, a collection of random perturbations is applied to the foreground regions before they are fed into the model. These perturbations include horizontal flipping, affine transformations, color jitter, Gaussian blur, and random erasing. These perturbations reduce the likelihood of the model overfitting to specific pixels or spatial patterns and promote the generation of more generalized representations. Moreover, since SCG relies on contrastive generation, increasing foreground variability through PAE also strengthens background learning, ultimately improving both the diversity and quality of the synthetic samples.

\subsubsection{Overall DFQ Pipeline}

The complete quantization pipeline of D4C consists of two stages, which are formally described in Algorithm~\ref{Algorithm_D4C}.

The first stage involves generating images using the pre-trained floating-point model. This process begins by initializing a batch of synthetic images with Gaussian noise and defining foreground and background regions via randomly generated bounding boxes. Simultaneously, a batch of representative text prompts is prepared for PGSI, covering categories such as animals, daily objects, food, and natural scenes. Foregrounds, backgrounds, and text prompts are then encoded using the corresponding image and text encoders, and the SCG loss is computed accordingly. Notably, both the foreground-text contrast from PGSI and the foreground-background contrast from SCG are integrated into Eq.~\ref{Eq_SCG}, enabling efficient joint optimization within a single backward pass. To further stabilize the generation process, we also incorporate a total variation regularization term~\cite{Dreaming_to_Distill}, denoted as $\mathcal{L}_{\text{TV}}$, leading to the final synthesis objective:
\begin{equation}
\mathcal{L}_{\text{Syn}} = \mathcal{L}_{\text{SCG}} + 0.1 \cdot \mathcal{L}_{\text{TV}}.
\label{Eq_Syn_Objective}
\end{equation}

\noindent The synthetic images are iteratively optimized to minimize $\mathcal{L}_{\text{Syn}}$ until convergence. This pipeline enables D4C to generate pseudo images that are both semantically informative and structurally diverse, thus facilitating more effective quantization and improving model performance.

In the second stage, we adopt an optimization-guided PTQ strategy to quantize the CLIP model. The synthesized images serve as calibration data to perform block- or layer-wise reconstruction on the image encoder. Meanwhile, the text prompts used in PGSI are repurposed to calibrate the text encoder via layer-wise reconstruction.

%% file: sec/4_Experiments.tex
\section{Experiments}

\subsection{Experiment and Implementation Details}

We evaluate the effectiveness of D4C on zero-shot classification across CIFAR-10, CIFAR-100~\cite{CIFAR}, and ImageNet-1K~\cite{ImageNet}. For the image encoder, we adopt RN50 and RN50x16 for CNNs, and VB32 and VB16 for ViTs. Pseudo images are first generated using D4C, followed by an optimization-based reconstruction PTQ process for model calibration. During pseudo-image generation, 180 representative prompts are curated for PGSI. These prompts also serve to calibrate and guide the optimization of the text encoder, adhering to the constraints of a DFQ scenario. We set the generation batch size to 16, adopt a learning rate of 0.01, and run the optimization for 3,000 iterations. The temperature parameter $\tau$ in the InfoNCE loss is set to 0.1. To ensure fair comparisons, we apply per-channel asymmetric quantization to weights and per-tensor asymmetric quantization to activations~\cite{QDrop}. However, for the QKV and Linear-1 projection layers in Transformers (including both ViTs and the text encoder), where activation quantization parameters can be folded into LayerNorm~\cite{RepQ-ViT}, we adopt per-channel quantization. We observe a performance bottleneck in the MLP layers of the text encoder; hence, we maintain 8-bit precision for these layers and apply per-channel quantization granularity. Addressing this challenge is left for future PTQ studies. Following common practice, the first convolution layer in both CNN and ViT image encoders is excluded from quantization~\cite{QDrop,PTQ4ViT}. For CNN-based encoders, we apply block-wise reconstruction~\cite{BRECQ} to optimize quantization parameters, whereas for Transformer-based encoders (image and text), we adopt a layer-wise reconstruction strategy~\cite{AdaRound}. We randomly sample 128 image inputs and 512 text inputs, initialize the quantization parameters using OMSE~\cite{OMSE}, and perform reconstruction for 20,000 iterations to ensure convergence, with learning rates set to $4.0\mathrm{e}{-}5$ for the image encoder and $4.0\mathrm{e}{-}6$ for the text encoder, respectively.

\begin{table}[htbp]
\small
\centering
\caption{Ablation study results evaluating the contribution of each component under the W6A6 quantization setting.}
\label{Table_Ablation_Comp}
\begin{tabular}{c|ccc|cc}
\toprule
\textbf{Dataset} & \textbf{PGSI} & \textbf{SCG} & \textbf{PAE} & \textbf{RN50} & \textbf{VB32} \\
\midrule
\multirow{5}{*}{\textbf{CIFAR-10}} 
 & & & & 58.9 & 19.1 \\
 & $\checkmark$ & & & 61.0 & 42.1 \\
 & $\checkmark$ & $\checkmark$ & & 67.2 & 63.2 \\
 & $\checkmark$ & & $\checkmark$ & 71.0 & 71.7 \\
 & $\checkmark$ & $\checkmark$ & $\checkmark$ & \textbf{73.4} & \textbf{75.5} \\
\midrule
\multirow{5}{*}{\textbf{CIFAR-100}} 
 & & & & 25.7 & 3.0 \\
 & $\checkmark$ & & & 29.8 & 16.6 \\
 & $\checkmark$ & $\checkmark$ & & 34.4 & 29.7 \\
 & $\checkmark$ & & $\checkmark$ & 34.8 & 41.4 \\
 & $\checkmark$ & $\checkmark$ & $\checkmark$ & \textbf{36.4} & \textbf{45.8} \\
 \midrule
\multirow{5}{*}{\textbf{ImageNet-1K}} 
 & & & & 39.5 & 15.7 \\
 & $\checkmark$ & & & 46.0 & 40.0 \\
 & $\checkmark$ & $\checkmark$ & & 46.5 & 45.6 \\
 & $\checkmark$ & & $\checkmark$ & 46.7 & 44.2 \\
 & $\checkmark$ & $\checkmark$ & $\checkmark$ & \textbf{47.9} & \textbf{46.1} \\
\bottomrule
\end{tabular}
\end{table}

\subsection{Experimental Results}

We evaluate D4C against three baselines: real samples, Gaussian noise, and existing DFQ methods, under W4A8, W6A6, and W8A8 quantization configurations. For the \textbf{Real} setting, image encoders are calibrated and optimized using randomly selected real images from the respective datasets. In the \textbf{Gaussian} setting, the initialized Gaussian noise is directly used for quantization. For the \textbf{BNS\&PSE} baseline, BNS~\cite{ZeroQ} is used to generate synthetic samples for CNN-based encoders, while PSE~\cite{PSAQ} is adopted for ViT-based encoders. These synthetic samples are then used to calibrate and quantize the corresponding models. The quantization results are summarized in Table~\ref{Table_Exp_Result}. Our proposed method, D4C, consistently outperforms all baselines, including the widely used BNS and PSE methods for CNN and ViT encoders, respectively. For instance, under W4A8 quantization, D4C improves the Top-1 accuracy of RN50/VB32 by 26.5/54.6\%, 6.8/39.0\%, and 17.9/32.5\% on CIFAR-10, CIFAR-100, and ImageNet-1K, respectively, when compared to Gaussian noise. It also achieves gains of 12.4/18.9\%, 6.8/19.7\%, and 1.4/5.7\% over the BNS\&PSE baseline. The results highlight the capability of D4C to generate high-quality synthetic samples with both semantic richness and diversity, positioning it as a compelling new baseline for DFQ on CLIP.

\subsection{Ablation Studies}

\subsubsection{Ablation of Components}

To evaluate the contribution of each component in D4C, we conduct ablation experiments on CIFAR-10, CIFAR-100, and ImageNet-1K using RN50 and VB32 under the W6A6 quantization configuration, as shown in Fig.~\ref{Table_Ablation_Comp}. Since PGSI forms the core of our approach, while SCG and PAE are optional enhancements, we additionally report results for PGSI combined with either SCG or PAE. The results reveal that each component contributes meaningfully to the final quantization performance, with accuracy improving incrementally as more modules are included. The best performance is achieved when all three components are applied together. These findings hold consistently across both CNN- and ViT-based encoders, underscoring the robustness and generalizability of D4C for CLIP.

\subsubsection{Ablation of PAE Perturbation}

To assess the individual effectiveness of each perturbation technique in PAE, we conduct an ablation study to evaluate their contributions to quantization performance. The PAE module incorporates five types of perturbations: horizontal flipping, affine transformation, color jitter, Gaussian blur, and random erasing, denoted as H, A, C, G, and R, respectively, in Table~\ref{Table_Ablation_PAE}. The results indicate that each perturbation contributes positively to the final quantization outcome, with the best performance achieved when all five are applied jointly. This validates the PAE module and highlights the complementary nature of these perturbations.

\begin{table}[htbp]
\small
\centering
\caption{Ablation study results of each perturbation technique in PAE under the W6A6 quantization setting.}
\label{Table_Ablation_PAE}
\begin{tabular}{c|ccccc|cc}
\toprule
\textbf{Dataset} & \textbf{H} & \textbf{A} & \textbf{C} & \textbf{G} & \textbf{R} & \textbf{RN50} & \textbf{VB32} \\
\midrule
\multirow{7}{*}{\textbf{CIFAR-10}} 
 & & & & & & 71.0 & 71.7 \\
 & $\checkmark$ & & & & & 71.2 & 74.0 \\
 & & $\checkmark$ & & & & 73.1 & 75.1 \\
 & & & $\checkmark$ & & & 71.4 & 72.7 \\
 & & & & $\checkmark$ & & 71.1 & 73.0 \\
 & & & & & $\checkmark$ & 71.2 & 72.4 \\
 & $\checkmark$ & $\checkmark$ & $\checkmark$ & $\checkmark$ & $\checkmark$ & \textbf{73.4} & \textbf{75.5} \\
 \midrule
\multirow{7}{*}{\textbf{ImageNet-1K}} 
 & & & & & & 46.7 & 44.2 \\
 & $\checkmark$ & & & & & 47.8 & 45.0 \\
 & & $\checkmark$ & & & & 46.8 & 45.9 \\
 & & & $\checkmark$ & & & 46.8 & 45.6 \\
 & & & & $\checkmark$ & & 47.4 & 45.8 \\
 & & & & & $\checkmark$ & 46.8 & 44.7 \\
 & $\checkmark$ & $\checkmark$ & $\checkmark$ & $\checkmark$ & $\checkmark$ & \textbf{47.9} & \textbf{46.1} \\
\bottomrule
\end{tabular}
\end{table}

\subsection{Synthetic Sample Visualization}

\begin{figure}[htbp]
    \centering
    \includegraphics[width=0.9\columnwidth]{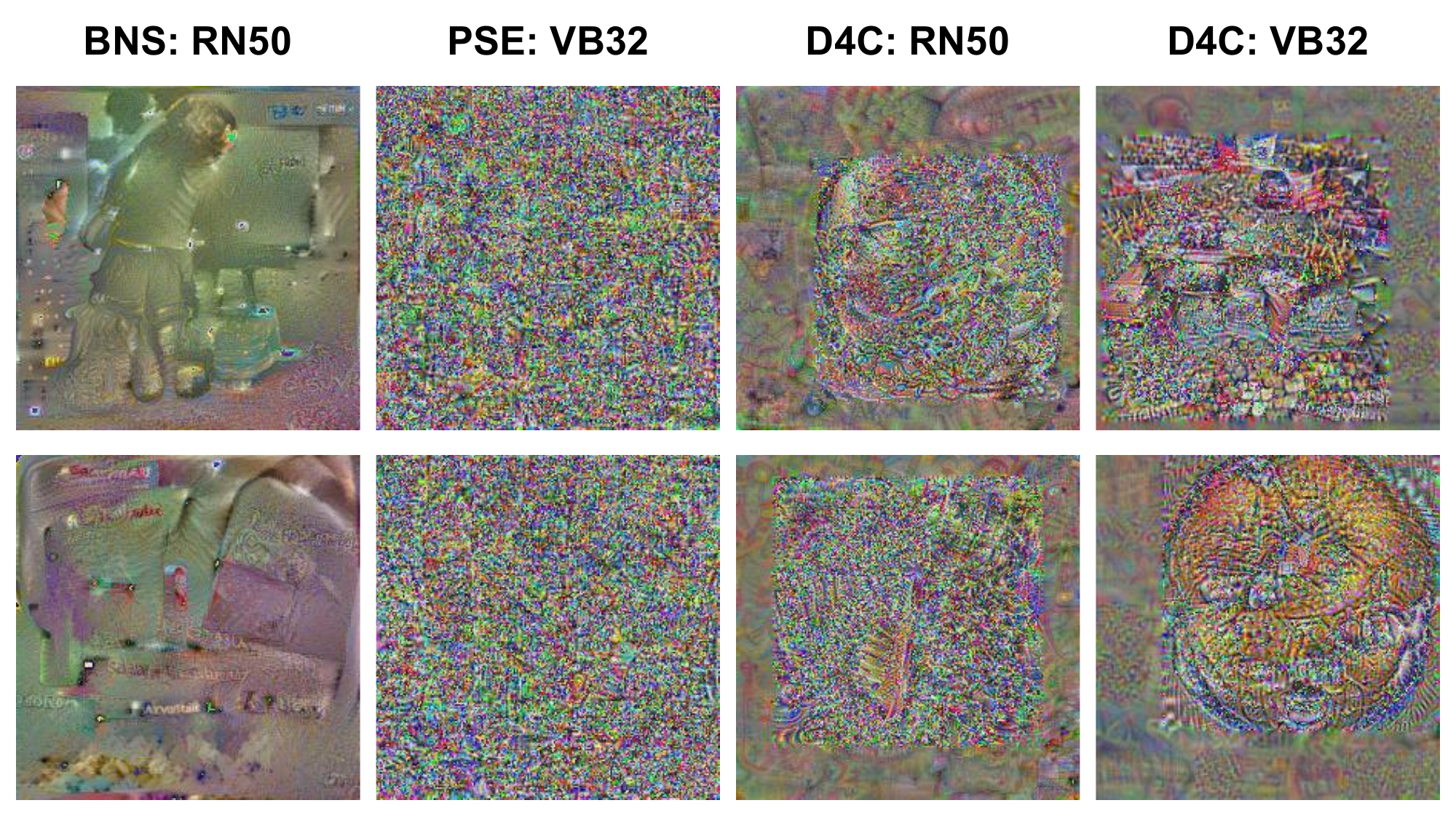}
    \caption{Visualization of synthetic samples generated by BNS and PSE (left) versus our proposed D4C framework (right) under both RN50 and VB32 encoders.}
    \label{Fig_Visualization}
\end{figure}

To qualitatively assess the effectiveness of our D4C framework in generating meaningful and structurally diverse pseudo images, we visualize the synthetic samples produced by different methods in Fig.~\ref{Fig_Visualization}. For RN50, samples generated via BNS loss tend to exhibit blurry and textureless patterns. Similarly, PSE-based synthesis for VB32 produces noisy and homogeneous textures with little discernible structure. In contrast, D4C generates visually more coherent and interpretable samples across both RN50 and VB32 encoders. These synthetic images exhibit clear semantic attributes (\emph{e.g.}, shapes and colors) and rich spatial structures resembling foreground-background compositions.

\subsection{Training Cost and Storage Saving}

Table~\ref{Table_Time_Cost} compares the training costs of various DFQ methods across different network architectures on ImageNet-1K. All experiments were conducted by generating 128 images on an RTX A6000 GPU (48 GB memory). The batch size was set to 16 for all methods, except PSE on VB16, which used a batch size of 8 due to memory limitations. As shown, D4C consistently delivers comparable or better efficiency than other baselines. For CNNs, D4C completes training in 1,623 s (RN50) and 12,491 s (RN50x16), slightly exceeding but remaining close to BNS (1,280 s and 9,515 s). In contrast, for ViTs, D4C demonstrates significant gains over PSE, reducing training time to 1,434 s (VB32) and 5,346 s (VB16), compared with 3,488 s and 44,398 s for PSE. Moreover, D4C efficiently supports both CNNs and ViTs, highlighting its strong generalization and practical utility.

\begin{table}[thbp]
\small
\centering
\caption{Training cost analysis for synthesizing 128 images.}
\label{Table_Time_Cost}
\begin{tabular}{c|cccc}
\toprule
 \textbf{Method} & \textbf{RN50} & \textbf{RN50x16} & \textbf{VB32} & \textbf{VB16} \\
\midrule
\textbf{BNS\&PSE} & \textbf{1,280} s & \textbf{9,515} s & 3,488 s  &  44,398 s \\
\textbf{D4C} & 1,623 s & 12,491 s & \textbf{1,434} s & \textbf{5,346} s \\
\bottomrule
\end{tabular}
\end{table}

We also evaluate the storage reduction and computational acceleration of our D4C-quantized model by measuring model size and estimating FLOPs. Compared to the FP32 baseline, our W4A8 quantized CLIP model achieves $5.63\times$ and $4.16\times$ storage reduction and speedup for RN50x16, and $5.45\times$ and $4.03\times$ for VB16, respectively. Even when the MLP layers in the text encoder remain at 8-bit precision, the additional storage and latency overheads are only 28\% and 13\% relative to full W4A8 quantization, while still maintaining an impressive overall compression ratio.

%% file: sec/5_Limitations.tex
\section{Limitations}

We highlight several limitations of D4C to suggest future directions. First, despite notable improvements over prior DFQ methods for CNNs and ViTs, a performance gap persists between synthetic and real samples, indicating room for further enhancement. Second, the quantization bottlenecks for CLIP remain underexplored, indicating a need for tailored quantization strategies.

%% file: sec/6_Conclusion.tex
\section{Conclusion}

In this work, we propose D4C, the first DFQ framework specifically designed for CLIP. D4C integrates three core components: PGSI, SCG, and PAE, to jointly mitigate the challenges of insufficient semantic richness and limited intra-image diversity in synthetic sample generation. Extensive experiments demonstrate that D4C consistently outperforms existing DFQ methods across various encoder architectures and quantization settings, underscoring the unique challenges of DFQ for CLIP and establishing D4C as a strong foundation for future research in this area.